\documentclass[lettersize,journal]{IEEEtran}
\usepackage{amsmath,amsfonts}
\usepackage{algorithmic}
\usepackage{algorithm}
\usepackage{array}
\usepackage[caption=false,font=normalsize,labelfont=sf,textfont=sf]{subfig}
\usepackage{textcomp}
\usepackage{stfloats}
\usepackage{url}
\usepackage{verbatim}
\usepackage{graphicx}
\usepackage{cite}
\usepackage{multirow}
\begin{document}

\title{Interpretable Neural Networks with Random Constructive Algorithm}

\author{Jing Nan, Wei Dai,~\IEEEmembership{Senior Member,~IEEE}



\thanks{}}

\markboth{IEEE}%
{Shell \MakeLowercase{\textit{et al.}}: Interpretable Neural Networks with Random Constructive Algorithm}


\maketitle

\begin{abstract}
This paper introduces an Interpretable Neural Network (INN) incorporating spatial information to tackle the opaque parameterization process of random weighted neural networks. The INN leverages spatial information to elucidate the connection between parameters and network residuals. Furthermore, it devises a geometric relationship strategy using a pool of candidate nodes and established relationships to select node parameters conducive to network convergence. Additionally, a lightweight version of INN tailored for large-scale data modeling tasks is proposed. The paper also showcases the infinite approximation property of INN. Experimental findings on various benchmark datasets and real-world industrial cases demonstrate INN's superiority over other neural networks of the same type in terms of modeling speed, accuracy, and network structure.
\end{abstract}


\section{Introduction}
\IEEEPARstart{N}{eural} networks, are different from traditional model-based methods in that they excel at extracting intrINNte patterns from complex data. Therefore, it is widely used in various research areas\cite{ref1,ref2,ref3}. Among them, Deep Neural Networks (DNNs) and Flat Neural Networks (FNNs) are the most prominent.DNNs achieve end-to-end learning by combining fine-tuning \cite{ref4}, which endows them with superior expressive and generalisation capabilities. However, DNNs are time-consuming to train. Recently, there has been a surge of interest in stochastic weighted neural networks (RWNNs), which are a typINNl FNN with universal approximation capabilities\cite{ref5,ref6,ref7,ref8}. RWNNs are run through a two-step training mode: firstly, randomly assigning the hidden parameters, and then evaluating the output weights through linear equations.

Random Weighted Neural Networks (RWNNs) have proven advantages, but their network structure is still relatively poorly suited for modelling tasks.  A network structure that is excessively large may lead to poor generalization, while one that is overly small may result in inadequate learning capacity. Constructive algorithms typically begin with a modest network structure, often just a single hidden node, and gradually expand it by incrementally adding new hidden nodes until the desired performance criteria are achieved \cite{ref9}. Consequently, these algorithms tend to provide more conservative network structures for modeling tasks \cite{ref10}. As a result, the constructive variant of RWNNs, IRWNNs, has been effectively employed in various data modeling endeavors \cite{ref11, ref12, ref13}.

In the domain of probability theory, the conventional approach of randomly generating hidden parameters may not always suffice for achieving optimal performance in IRWNNs. This raises a fundamental question: What defines suitable hidden parameters for IRWNNs? Through a thorough algebraic exploration of multidimensional nonlinear functions, researchers \cite{ref14} have demonstrated that the relationship between input samples and hidden parameters can be aptly characterized by nonlinear weight equations. Moreover, insights from \cite{ref15} have unveiled the presence of a supervisory mechanism between hidden parameters and input samples, thereby enhancing network performance. Building upon these foundational insights, \cite{ref16} introduced a constructive algorithm incorporating a supervisory mechanism to effectively allocate hidden parameters within a dynamic interval. Additionally, \cite{ref17} proposed a methodology for generating hidden parameters by analyzing the range of input samples and activation functions. More recently, \cite{ref18} introduced RWNs with compact incremental inequality constraints, referred to as CIRW, aimed at refining the quality of hidden parameters. Despite these advancements, the precise mechanisms through which hidden parameters fulfill their objectives remain largely unexplored. Consequently, visualizing the influence of each hidden parameter on residual error (indicative of network performance) remains a daunting challenge. Currently, enhancing the interpretability of predicted behaviors of NNs is gaining increasing attention and significance in research \cite{ref19, ref20}. Hence, further exploration into interpretable constructive algorithms is deemed crucial and warranted for advancing the field of RWNs.

This paper introduces an interpretable neural network. The primary contributions are outlined as follows:

1) Utilizing the spatial relationship between hidden parameters and network error, this paper devises an interpretable spatial information constraint to guide the assignment of randomized hidden parameters.

2) In two implementations of the algorithms, INN and IN+, which utilise different methods of calculating the output weights of the network, a pool of nodes is used to systematically search for hidden parameters, thus improving the quality of the hidden nodes.

Section II provides a brief overview of RWNNs. In section III, we introduce our proposed INN in detail. Section IV presents the experimental evaluation. Finally, Section V concludes the article, summarising the findings and discussing potential future directions.

\section{PROBLEM ANALYSIS}
\subsection{RWNN and Constructive Algorithms}
In RWNN, all hidden parameters are randomly assigned from a predetermined interval and remain fixed throughout the training process. The evaluation of output weights is accomplished by solving a system of linear equations. The RWNN are expounded as follows.

For $f:{R^d} \to {R^m}$, the RWNNs with $L$ hidden nodes is ${f_L} = H\beta $, where $H = \left[ {{g_1}\left( {\omega _1^{\rm{T}} \cdot x + {b_1}} \right), \cdots ,{g_L}\left( {\omega _L^{\rm{T}} \cdot x + {b_L}} \right)} \right]$, T denotes matrix transpose, $x$ is the input sample, ${\omega _j}$ and ${b_j}$ are the input weights and biases of the $j$-th hidden node, respectively. ${g_j}$ denotes the nonlinear activation function. ${\beta}$ are evaluated by $\beta  = {H^\dag }{f_L}$, where $\beta  = \left[ {{\beta _1},{\beta _2},...,{\beta _L}} \right]_{}^{\rm{T}}$.

Constructive algorithms, owing to their incremental construction approach, tend to discover the minimal network structure. Consequently, these algorithms have been adapted to RWNNs, leading to the introduction of IRWNNs. In particular, if an IRWNN with $L - 1$ hidden nodes fails to meet the termination condition, a new hidden node is generated through the following two steps:

1) The input weights  and bias  are randomly generated from the fixed interval. Then, the output vector ${g_L}$, which is determined by $\Delta  = \frac{{{{\left\langle {{e_{L - 1}}} \right\rangle }^2}}}{{{{\left\| {{g_L}} \right\|}^2}}}$. ${f_{L - 1}}$ is the output of IRWNNs.

2) ${\beta _L}$ is the output weight of the $L$-th hidden node.

If ${e_L} = f - {f_L}$ does not meet the predefined residual error, then a new hidden node should be added until the predefined residual error is reached.

\section{Interpretable neural network}
This section establishes the interpretable spatial information constraint by leveraging the geometric relationship between the network error and the parameters. The universal approximation property of this constraint is ensured by integrating the residual error. Furthermore, a node pool strategy is adopted to acquire hidden parameters that promote convergence. Lastly, two distinct algorithm implementations are introduced: INN and IN+.
\vspace{-1em}
\subsection{Interpretable Spatial Information Constraint}

\noindent \quad {\bf{Theorem 1:}} Suppose that span($\Gamma $) is dense in ${L^2}$ and $\forall g \in \Gamma $, $0 < \left\| g \right\| < v$ for some $v \in R$.  If ${g_L}$ is randomly generated under interpretable spatial information constraint
\begin{equation}
\setlength{\abovedisplayskip}{2pt}
\setlength{\belowdisplayskip}{2pt}
{\cos}{\theta _{L}} \ge {\gamma _L}\left\langle {{e_{L}},{e_{L}}} \right\rangle
\end{equation}

${\beta _L}$ are evaluated by ${\beta _L}{\rm{ = }}\frac{{\left\langle {{e_{L - 1}},{g_L}} \right\rangle }}{{{{\left\| {{g_L}} \right\|}^2}}}$. Then, we have ${\lim _{L \to  + \infty }}\left\| {{e_L}} \right\| = 0$.

\noindent \quad {\bf{Proof:}}
It's easy to know that $\left\| {{e_L}} \right\|$ is monotonically decreasing as $L \to \infty $.

It follows from Eq. (1) that

\begin{equation}
\label{deqn_ex1a}
\begin{array}{l}
{\rm{   }}{\left\| {{e_L}} \right\|^2} - \tau {\left\| {{e_{L - 1}}} \right\|^2}\\
 = \sum\limits_{q = 1}^m {\left\langle {{e_{L - 1,q}} - {\beta _{L,q}}{g_L},{e_{L - 1,q}} - {\beta _{L,q}}{g_L}} \right\rangle } \\
{\rm{     }} \quad - \sum\limits_{q = 1}^m {\tau \left\langle {{e_{L - 1,q}},{e_{L - 1,q}}} \right\rangle } \\
 = {\gamma _L}{\left\| {{e_{L - 1}}} \right\|^2} - \frac{{{{\left\langle {{e_{L - 1}},{g_L}} \right\rangle }^2}}}{{{{\left\| {{g_L}} \right\|}^2}}}
\end{array}
\end{equation}

Then, we have
\begin{equation}
\setlength{\abovedisplayskip}{2pt}
\setlength{\belowdisplayskip}{2pt}
\begin{array}{l}
\quad {\rm{   }}\left\langle {{e_L},{g_L}} \right\rangle \\
{\rm{ = }}\left\langle {{e_{L - 1}},{g_L}} \right\rangle  - \frac{{\left\langle {{e_{L - 1}},{g_L}} \right\rangle }}{{{{\left\| {{g_L}} \right\|}^2}}}\left\langle {{g_L},{g_L}} \right\rangle \\
 = 0
\end{array}
\end{equation}

Then, we have
\begin{equation}
\setlength{\abovedisplayskip}{2pt}
\setlength{\belowdisplayskip}{2pt}
\begin{array}{l}
{\rm{    \quad   }}\sum\limits_{j = 1}^{L - 1} {{\beta _j}\left\langle {{e_{L - 1}},{g_j}} \right\rangle } \\
{\rm{   }} = \left\langle {{e_{L - 1}},\sum\limits_{j = 1}^{L - 1} {{\beta _j}{g_j}} } \right\rangle \\
{\rm{   }} = {\left\| {{e_{L - 1}}} \right\|^2}
\end{array}
\end{equation}
where ${f_{L - 1}}$ is orthogonal to ${e_{L - 1}}$. Thus, we have
\begin{equation}
\begin{array}{l}
\quad {\rm{       }}\exists {\beta _j}\left\langle {{e_{L - 1}},{g_j}} \right\rangle  \ge \frac{{{{\left\| {{e_{L - 1}}} \right\|}^2}}}{{L - 1}}\\
{\rm{ }} =  > \frac{{\left| {\left\langle {{e_{L - 1}},{g_j}} \right\rangle } \right|}}{{\left\| {{g_j}} \right\|}} \ge \frac{{{{\left\| {{e_{L - 1}}} \right\|}^2}}}{{\left( {L - 1} \right){\beta _j}\left\| {{g_j}} \right\|}}
\end{array}
\end{equation}
where $\frac{{{{\left\| {{e_{L - 1}}} \right\|}^2}}}{{L - 1}}$is the average of ${\left\| {{e_{L - 1}}} \right\|^2}$.

According to the Fig. 1 and Eq. (5), we have
\begin{equation}
\begin{array}{l}
\frac{{\left| {\left\langle {{g_j},{e_{L - 1}}} \right\rangle } \right|}}{{\left\| {{g_j}} \right\|}} = \left( {\left\| {{e_{L - 1}}} \right\|\cos {\theta _{L - 1}}} \right)
\end{array}
\end{equation}

Then, we have
\begin{equation}
{\cos}{\theta _{L }} \ge \varphi {\left\| {{e_{L}}} \right\|^2}
\end{equation}

To ensure that Equation (9) holds, ${\gamma _L}$ is designed as a dynamic value.
\begin{equation}
\varphi  \ge {\gamma _L}
\end{equation}

It follows from Eq. (1), (3), (8), and (9) that
\begin{equation}
\begin{array}{l}
\quad {\rm{   }}{\left\| {{e_L}} \right\|^2} - \tau {\left\| {{e_{L - 1}}} \right\|^2}\\
 \le {\gamma _L}{\left\| {{e_{L - 1}}} \right\|^2} - {\cos ^2}{\theta _{L - 1}}\\
 \le 0
\end{array}
\end{equation}

Then, we have that ${\lim _{L \to  + \infty }}\left\| {{e_L}} \right\| = 0$.

\subsection{Node Pooling}
While the interpretable spatial information constraint guarantees the UAP of the INN, ${g_L}$ are initially generated randomly in a single step. This approach may not facilitate rapid reduction of the network residual. Consequently, to optimize Eq. (1), the node pool is employed as follows:
\begin{equation}
{\left( {{{\cos }^2}{\theta _L}} \right)_{\max }} \ge {\gamma _L}\left\langle {{e_{L - 1}},{e_{L - 1}}} \right\rangle
\end{equation}

\begin{figure}[!h]
\centering
\includegraphics[width=3in]{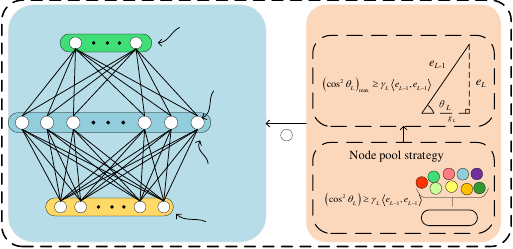}
\caption{Network structure of INN.}
\label{fig_2}
\end{figure}

\begin{figure}[!h]
\centering
\includegraphics[width=2in]{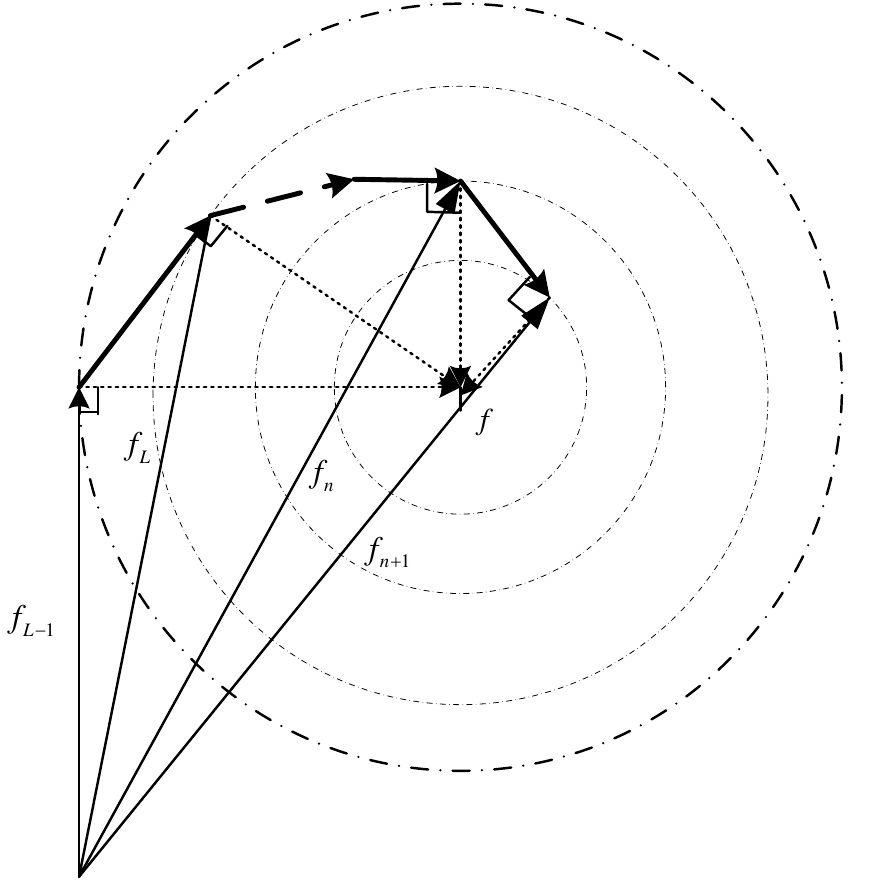}
\caption{construction process of INN.}
\label{fig_3}
\end{figure}
\subsection{Algorithm steps}
Here in this paper, two algorithmic implementations, called INN and IN+, are proposed. Figures 1 and 2 illustrate the network structure and construction process of IC, respectively.
\subsubsection{INN}
For a target function $f:{R^d} \to {R^m}$,  let's assume that an IC with ${L-1}$ hidden nodes has been constructed, i.e., ${f_{L - 1}} = \sum\nolimits_{j = 1}^{L-1} {{\beta _j}{g_j}\left( {\omega _j^{\rm{T}} \cdot x + {b_j}} \right)} $. If the generated ${g_L}$ makes the interpretable geometric information constraint Eq. (11) hold, and the output weights are given by
%
\begin{equation}
\label{deqn_ex1a}
\beta  = {H^\dag }{f_L}
\end{equation}

\noindent \quad {\bf{Remark 3:}} 
Equation 12 is time-consuming when modelling large samples, for this reason we propose a lightweight version, based on Greville theory\cite{ref21}, \cite{ref22}, \cite{ref23}. 
\subsubsection{IN+}
Let${H_L}{\rm{ = }}\left[ {{H_{L{\rm{ - }}1}}{\rm{ }}{g_L}} \right]$, then, we have that according to Gravell's iteration theory
\begin{equation}
H_L^\dag {\rm{ = }}\left[ {\begin{array}{*{20}{c}}
{H_{L - 1}^\dag  - {d_L}b_L^{\rm{T}}}\\
{b_L^{\rm{T}}}
\end{array}} \right]
\end{equation}

Then, the output weights can be derived as
\begin{equation}
\beta {\rm{ = }}\left[ {\begin{array}{*{20}{c}}
{{\beta ^{previous}} - {d_L}b_L^{\rm{T}}f}\\
{b_L^{\rm{T}}f}
\end{array}} \right]
\end{equation}
\section{EXPERIMENTAL RESULTS}
In this section, we present a comparative analysis of the proposed methods: INN, IN+ and other stochastic algorithm comparisons. We leverage six benchmark datasets, one actual mill dataset, and one gesture recognition dataset for our evaluation. The details of these datasets are summarized in Table I. Additionally, Table II provides an overview of each algorithm's information, including experimental parameters used on these datasets.

\begin{equation}
f\left( x \right) = \frac{1}{{({{(x - 0.3)}^2})}} + \frac{1}{{({{(x - 0.9)}^2} )}} - 6
\end{equation}

In MATLAB 2020a, the comparison experiments were performed on a personal computer. Each experiment was demonstrated after several repetitions and averaged.
\begin{table}[!t]
\caption{Description of Datasets\label{tab:table1}}
\centering
\begin{tabular}{ccccc}
\hline
Data & Test&Train & Various & Label\\
\hline
DB1 &  400 &2000 & 1 & 1\\
DB2 &  258 &2444 & 561 & 6\\
 DB3& 5330 & 7200 & 8 & 1\\
DB4 &  479 & 1120 &15 & 3\\
DB5 &  45 & 105 &4 & 3\\
 DB6& 2947 &7563 &  15 & 1\\
DB7 &  523 & 1237 &1 & 1\\
\hline
\end{tabular}
\end{table}
\begin{table}[!t]
\caption{Parameters\label{tab:table2}}
\centering
\begin{tabular}{ccccccc}
\hline
 \multirow{2}{*}{Data} & \multicolumn{2}{c}{ IRW}& \multicolumn{2}{c}{CIR/INN} & \multirow{2}{*}{$\ell $}  & \multirow{2}{*}{${L_{\max }}$}  \\
 \cline{2-5}
  &  $\lambda $ & ${T_{\max }}$ & $\zeta $& ${T_{\max }}$ & &\\
\hline
DB1 & 15& & 1:1:200 &  10 & 0.05 & 30\\
DB2 & 1.5& & 5:1:50 &  5 & 0.05 & 50\\
DB3 & 2& & 1:0.1:3 & 1  & 0.05 & 800\\
DB4 &3& 1 &1:1:5 & 30  & 0.05 & 30\\
DB5 & 50& & 1:5:200 &  15 & 0.05 & 10\\
DB6 & 0.5& & 1.5:0.1:7 & 5 & 0.05 & 200\\
DB7& 2& & 100:10:200 &  16 & 0.05 & 100\\
\hline
\end{tabular}
\end{table}
\begin{figure}[!t]
\centering
\includegraphics[width=2.5in]{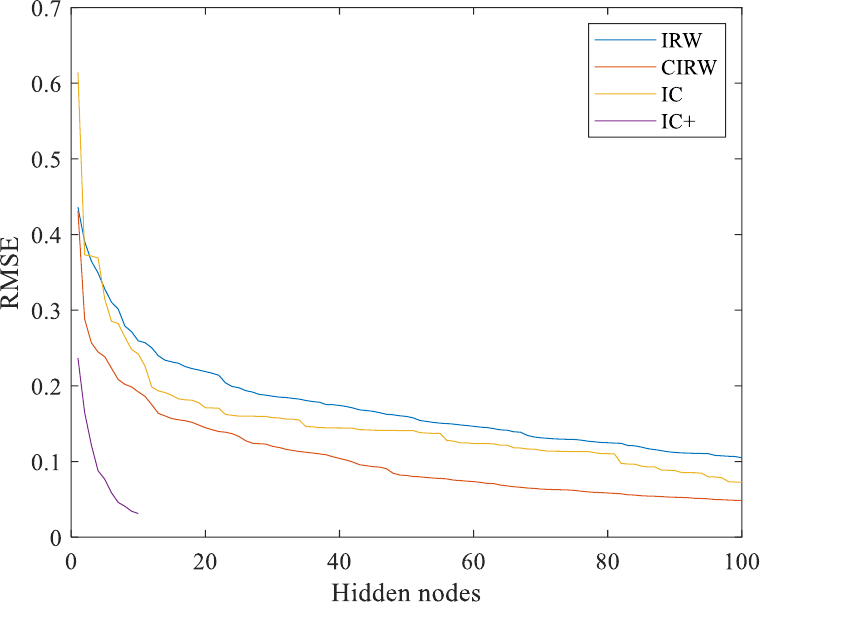}
\caption{Convergence of RMSE.}
\label{fig_4}
\end{figure}
\begin{figure}[!t]
\centering
\includegraphics[width=2.5in]{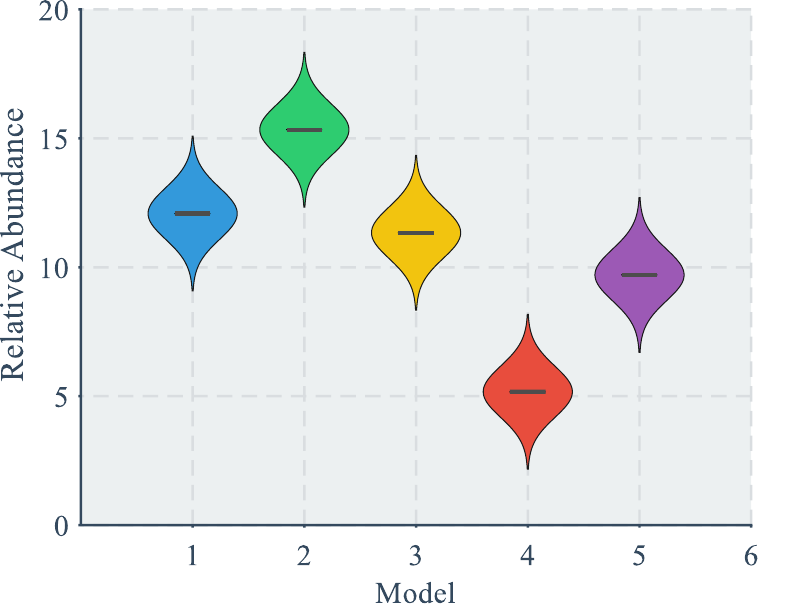}
\caption{KDF for four algorithms.}
\label{fig_5}
\end{figure}
\subsection{Results}
\subsubsection{Function Approximation Dataset}

Figure 4 depicts the RMSE performance of the four algorithms across the function dataset. It is evident that both the proposed method and CIRW exhibit rapid convergence to the desired RMSE with a minimal number of hidden nodes. Furthermore, both variants of INN require fewer hidden nodes to achieve the predefined objective, highlighting the compactness advantage of the proposed INN. Transitioning to Figure 5, it presents the KDF of the estimation error. Notably, the KDF of INN closely aligns with the distribution of real data, indicating superior prediction capability compared to other algorithms.

In Figure 6, we explore the impact of different parameters, denoted as $\lambda$, on the KDF of IC. Results indicate that varying parameters yield varying effects on the KDF performance of IC, suggesting that maintaining $\lambda$ as a fixed value may not optimize KDF performance. This observation underscores the importance of thorough parameter tuning and optimization to achieve optimal model performance and predictive accuracy in the context of IC.

\begin{table*}[!t]
    \caption{Comparison with Four Algorithms in Terms of Time, Training RMSE and Testing RMSE on the benchmark datasets}
    \centering
\begin{tabular}{ccccccccccccc}
        \hline
\multirow{2}{*}{Dataset} & \multicolumn{12}{c}{Training time Training RMSE Testing RMSE}   \\
                         \cline{2-13}
                         & \multicolumn{3}{c}{IRW} & \multicolumn{3}{c}{CIR} & \multicolumn{3}{c}{IC}  & \multicolumn{3}{c}{IC+}\\
                         \hline
Compactiv                & 0.234s   & 0.283  & 0.257  & 1.215s   & 0.062  & 0.071 & 1.122s   & 0.062 & 0.072 & 0.580s   & 0.062 & 0.072\\
Concrete                 & 0.278s   & 0.253  & 0.256  & 1.09s    & 0.106  & 0.206 & 0.852s   & 0.095 & 0.206 & 0.281s   & 0.134 & 0.232\\
Winequality         & 0.120s   & 0.308  & 0.308  & 0.314s   & 0.150  & 0.159 & 0.287s   & 0.150 & 0.185 & 0.069s   & 0.120 & 0.185\\
HAR                    & 0.059s   & 0.160  & 0.239  & 0.08s   & 0.019  & 0.043 & 0.080s   & 0.017 & 0.042 & 0.022s   & 0.028 & 0.043\\
ORE                      & 56.401s   & 0.117  & 0.149  & 123.90s  & 0.016  & 0.055 & 123.177s & 0.314 & 0.036 & 40.259s  & 0.014 & 0.038\\
Segment                  & 0.318s   & 0.277  & 0.479  & 0.590s   & 0.199  & 0.218 & 0.536s   & 0.194 & 0.216 & 0.265s   & 0.193 & 0.226\\
        \hline
\end{tabular}
\end{table*}
\begin{figure}[!t]
\centering
\includegraphics[width=2.5in]{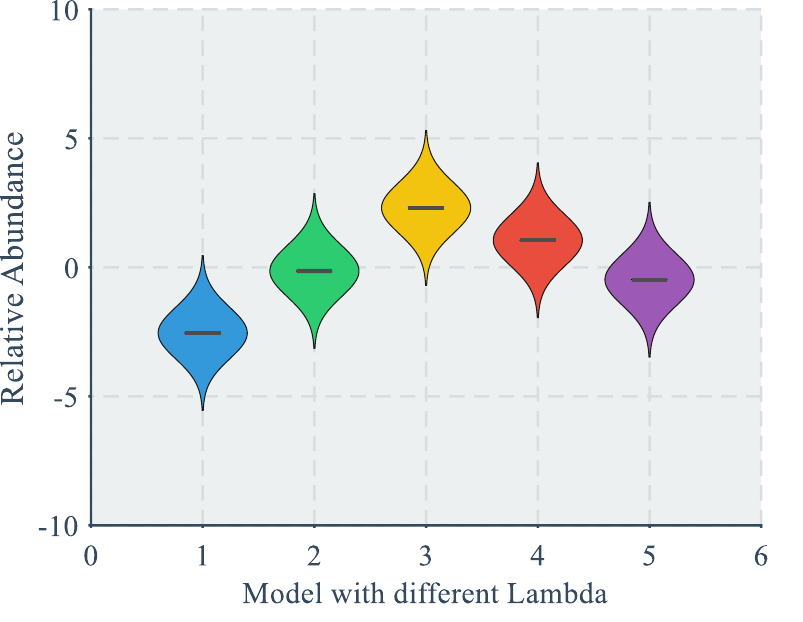}
\caption{Impact of $\lambda $ on KDF.}
\label{fig_6}
\end{figure}
\subsubsection{Datasets}

In this section, we delve into the performance analysis of the four algorithms across a spectrum of benchmark datasets. Table III offers insights into the elapsed time and RMSE of these algorithms on the respective datasets. A comprehensive examination of the data in Table III unveils a consistent trend: the training RMSE values of INN consistently outperform those of the other algorithms across all datasets. This consistent superiority suggests that INN possesses the capability to generate higher quality nodes compared to its counterparts. Furthermore, it is noteworthy to highlight that IN+ exhibits weaker performance in comparison. This discrepancy can be attributed to the heavy reliance of IN+ on the quality of the output weights of the initial hidden nodes. The subpar performance of IN+ underscores the importance of carefully considering the initialization strategy and optimization techniques employed in the model, as these factors significantly influence the overall performance and efficacy of the algorithm.

When adding the same number of hidden nodes, it becomes evident that the proposed INN requires less time compared to CIRW. Furthermore, across most datasets, IN+ demonstrates a significant reduction in training time compared to INN, thereby highlighting its superior efficiency. Moving forward to Figure 7, it visually depicts the impact of hidden node pooling on the performance of INN. The experimental findings suggest that both excessively large and excessively small node pools lead to an increase in RMSE. This observation underscores the crucial role of node pool size in influencing the network's efficiency and performance. Therefore, careful consideration of this parameter is imperative during the experimentation process, as highlighted in this study.
\begin{figure}[!t]
\centering
\includegraphics[width=2.5in]{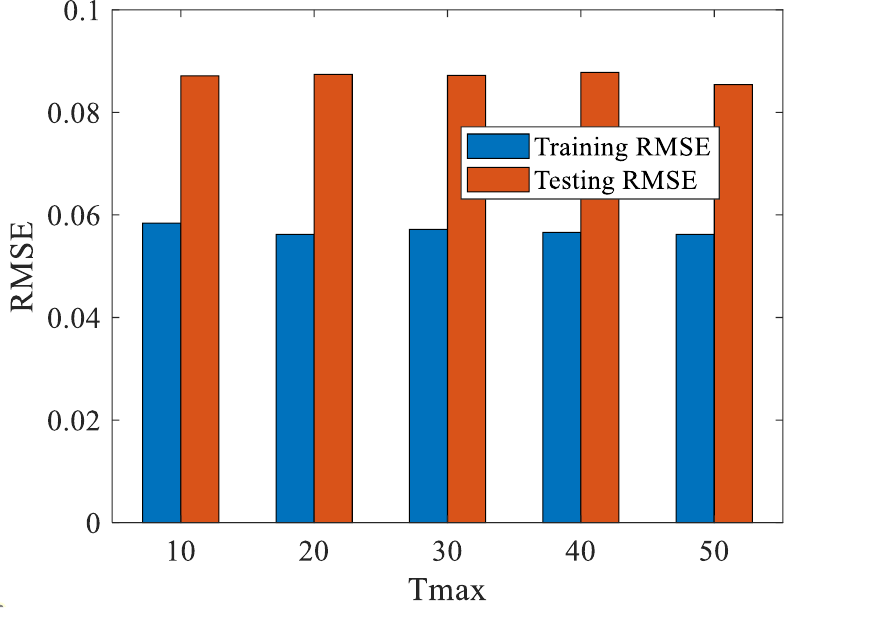}
\caption{Impact of node pool on RMSE.}
\label{fig_7}
\end{figure}
\subsection{HGR}
HGR finds applications across a spectrum of fields including Augmented Reality (AR), smart home automation, and human health monitoring \cite{ref24}. In this section, we focus on evaluating the performance of INN within the context of an HGR system. As depicted in Figure 8, the HGR system comprises two essential modules: hardware and software. For our evaluation, we leverage the software component to gather the HGR dataset. This dataset encompasses a total of 2536 samples \cite{ref25}. Through our analysis, we aim to elucidate the effectiveness and robustness of INN in the realm of hand gesture recognition, shedding light on its potential applications in various domains such as AR, smart home systems, and healthcare monitoring.

Translated with www.DeepL.com/Translator (free version)
\subsubsection{Comparison}
Figure 11 displays the Root Mean Square Error (RMSE) of INN in comparison to other methods on HGR. It is noteworthy that INN exhibits remarkable stability in terms of RMSE. Furthermore, the disparity between the maximum and minimum RMSE values of the other algorithms is notably larger, with differences of 0.15 and 0.02, respectively. Table IV provides a comprehensive overview of the experimental outcomes of the four methods applied to HGR. INN demonstrates a clear advantage in both training time and accuracy metrics. Through a thorough comparison and analysis of these results, it becomes evident that the proposed INN outperforms other algorithms in the HGR task, showcasing its efficacy and superiority in this domain.
\begin{figure}[!t]
\centering
\includegraphics[width=3.5in]{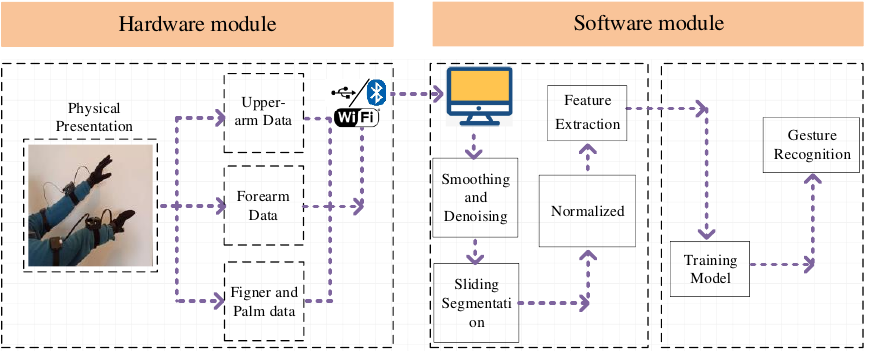}
\caption{Framework of HGR.}
\label{fig_8}
\end{figure}
\begin{figure}[!t]
\centering
\includegraphics[width=2.5in]{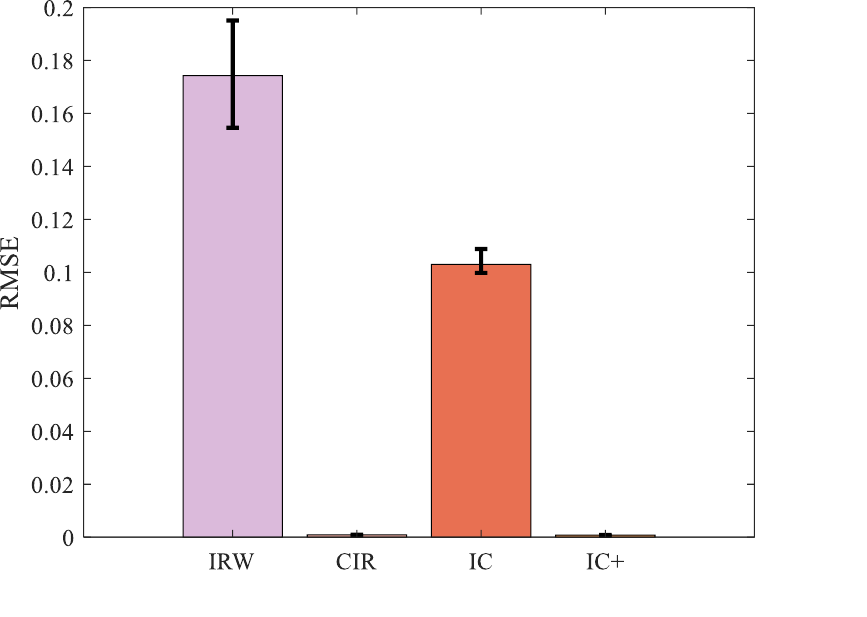}
\caption{RMSE of four algorithms.}
\label{fig_11}
\end{figure}
\begin{table}[!t]
\caption{Performance Comparison of Four Algorithms on HGR\label{tab:table1}}
\centering
\begin{tabular}{cccc}
\hline
Algorithms &time &Accuracy & Nodes\\
\hline
IRW & 20.37s & 82.43\% & 500\\
CIR & 41.63s &95.12\% & 500\\
INN & 40.79 & 96.10\% & 500\\
IN+ & 14.68s & 96.48\% & 500\\
\hline
\end{tabular}
\end{table}
\subsection{ORE}
Ore grinding, a crucial step in mineral processing, entails the complete dissociation of valuable ores. The underlying mechanism of ore grinding is highly intricate and multifaceted, rendering the development of a comprehensive mathematical model exceedingly difficult. Consequently, the imperative arises to devise a data-driven model that can proficiently monitor the grinding process. The primary aim of the ore grinding model is to discern and delineate the intricate mapping relationships between various variables involved in the grinding process. By accurately capturing these relationships, the model can provide valuable insights into optimizing the grinding process, enhancing efficiency, and maximizing the extraction of valuable minerals from ores.
\begin{equation}
{\rm{ORE}} = f\left( {{B_1},{B_2},{B_3},{A _1},{A _2}} \right)
\end{equation}
\subsubsection{Discussion}

Figure 14 illustrates the error Probability Density Functions (PDFs) of INN alongside its comparison methods. Notably, INN and IN+ exhibit a similar PDF curve owing to their comparable performance. It is evident that both INN and IN+ surpass the other methods as their curves closely resemble the normal distribution compared to alternative algorithms. This observation underscores IN's superior performance in Outlier Recognition and Elimination (ORE). Table V presents the experimental outcomes of the four methods applied to ORE. Despite INN's prolonged training duration, it excels in terms of accuracy. Through the experimental results and analysis, the proposed algorithm demonstrates commendable performance in terms of generalization and training time.
\begin{figure}[!t]
\centering
\includegraphics[width=3.5in]{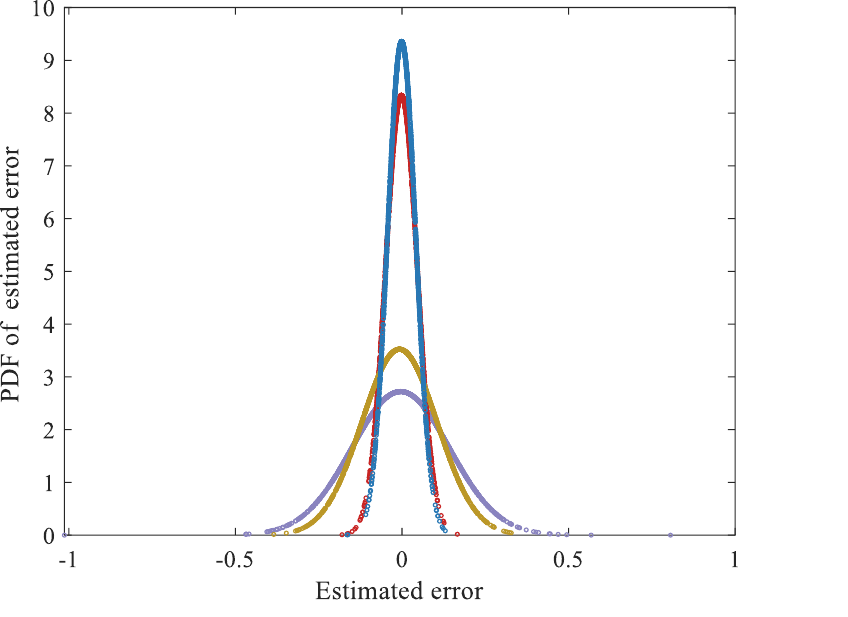}
\caption{PDF of all algorithms.}
\label{fig_14}
\end{figure}
\begin{table}[!t]
\caption{Performance Comparison of Four Algorithms on Ore\label{tab:table1}}
\centering
\begin{tabular}{cccc}
\hline
Algorithms &Training time &Accuracy & Nodes\\
\hline
IRW & 0.57s & 85.89\% & 100\\
CIR & 3.93s &95.53\% & 100\\
INN & 3.29s & 95.67\% & 100\\
IN+ & 1.91s & 95.81\% & 100\\
\hline
\end{tabular}
\end{table}
\section{Conclusion}
IN+ is a further extension of Interpretable Neural Networks (INN), aimed at enhancing both efficiency and performance. Unlike traditional INN, IN+ introduces new features and techniques to further optimize the network's behavior. Firstly, IN+ offers a more efficient method for generating nodes. Instead of the spatially constrained random node generation used in INN, IN+ employs smarter algorithms or optimization strategies to select and arrange nodes, thereby improving the training and inference efficiency of the network. Secondly, IN+ introduces more refined techniques for analyzing node influences. In addition to intuitively demonstrating the impact of each hidden node on network error, IN+ further analyzes the interactions and influences between nodes to provide a more comprehensive and detailed explanation.  Furthermore, IN+ may include customized optimizations tailored to specific tasks or datasets. By conducting in-depth analyses of the requirements of different tasks and adjusting the network structure, parameters, or algorithms accordingly, IN+ can achieve better performance. Finally, extensive experiments on various public and real-world datasets have been conducted to validate IN+. The experimental results demonstrate that compared to traditional INN or other methods, IN+ not only reduces computational overhead but also further enhances network performance and behavior. Overall, IN+ represents a significant advancement in the field of interpretable neural networks. It not only provides more effective explanatory capabilities but also achieves significant improvements in performance and efficiency, opening up new directions for the development of interpretable artificial intelligence.

\end{document}